# Multi-Angular Reflectance Anisotropy Observed from UAV Multispectral Imagery

Zhenqiang Qin
Geospatial Information College
Information Engineering University
Zhengzhou, China
18530399015@163.com

Chenguang Dai
Geospatial Information College
Information Engineering University
Zhengzhou, China
cgdai2008@163.com

Min Wang
Hebi Meteorological Bureau of Henan Province
Hebi, China
wangqhebi@163.com

Xian Li*
School of Electronics and Information Engineering
Harbin Institute of Technology
Harbin, China
xianli@hit.edu.cn

***Abstract*—UAV multispectral imagery naturally contains multi-angular observations due to low flight altitude and wide field-of-view imaging, which may introduce geometry-driven radiometric variability. This study proposes a geometry-aware multi-angular observation extraction workflow to quantify observation-geometry effects from a BRDF perspective. Specifically, camera intrinsics and extrinsics are refined via structure-from-motion (SFM), and homogeneous regions annotated on an orthomosaic are reprojected onto multiple raw sub-images acquired from different viewpoints. This enables joint extraction of multi-band reflectance and observation geometry parameters for the same ground targets under varying viewing directions. The extracted observations are further analyzed using band-wise polar visualization in the (VZA, RAA) domain. Results on a grassland target show clear reflectance anisotropy across ten bands, with red-edge and near-infrared bands exhibiting 119–137% variability between maximum and minimum reflectance, indicating non-negligible observation-geometry effects on radiometric consistency.**

***Keywords—UAV, multispectral, observation geometry, BRDF, multi-angular observation***

## I. INTRODUCTION

Unmanned aerial vehicles (UAVs) equipped with multispectral sensors have been widely used in remote sensing due to their high spatial resolution and flexibility. However, low flight altitude and wide field-of-view imaging systems naturally produce multi-angular observations even within a single flight, meaning that the same surface target can be observed under varying viewing geometries. Such geometry-driven directional variations in reflectance may degrade radiometric consistency and affect quantitative interpretation.

The angular dependence of surface reflectance is commonly described by the BRDF [1]. Although BRDF studies have been extensively conducted for satellite multi-angular remote sensing, UAV platforms have also demonstrated strong potential for high-resolution multi-angular reflectance measurements and anisotropy characterization [3]–[5], as well as high-precision BRDF modeling [2]. In addition, polarization and multi-angular reflectance measurements from UAVs further highlight the importance of geometry-dependent effects in close-range remote sensing [4]. Meanwhile, recent studies have shown the feasibility of extracting BRDF-related information from UAV multispectral observations and emphasized the value of robust geometric processing in UAV workflows [6]-[8].

Motivated by these studies, this work proposes a **geometry-aware multi-angular observation extraction workflow** to quantify observation-geometry effects from a BRDF perspective. SFM-refined camera parameters are used to reproject orthomosaic-annotated homogeneous regions onto multiple sub-images acquired from different viewpoints, enabling joint extraction of multi-band reflectance and observation geometry for the same ground targets under different views. Band-wise polar visualization in the view zenith angle, relative azimuth angle (VZA, RAA) domain is further employed to reveal geometry-driven reflectance patterns and provide an intuitive interpretation of multi-angular reflectance variability in UAV multispectral imagery.

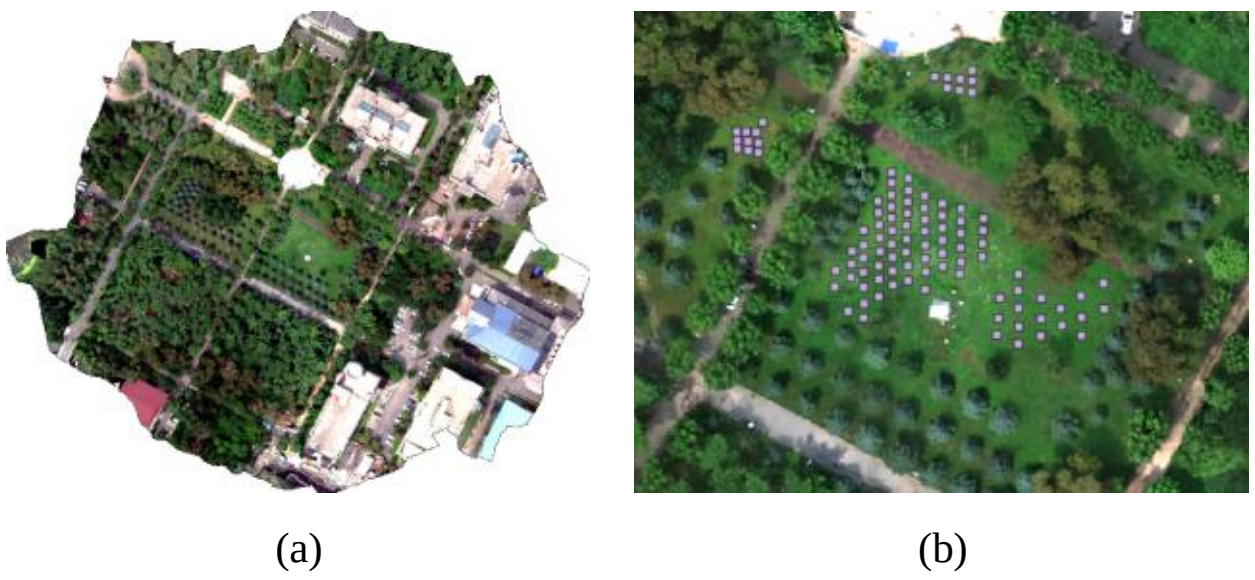

(a) (b)

Fig. 1. Orthomosaic of the study area on the HIT campus and selected homogeneous grassland ROIs.

## II. STUDY AREA AND DATA

### A. Study Area

The experiments were conducted on the Harbin Institute of Technology (HIT) campus, where a grassland area with relatively homogeneous surface conditions was selected as the case study, as shown in Fig.1. The site exhibits limited topographic variation, and the surface condition remained stable during the UAV acquisition, making it suitable for analyzing reflectance variations primarily driven by observation geometry.

An orthomosaic was generated from the UAV image to provide an overview of the scene. Based on the orthomosaic, multiple homogeneous grassland sub-regions were manually selected as regions of interest (ROIs) for multi-angular analysis. These ROIs exhibit good spectral continuity, which helps reduce the influence of land-cover heterogeneity and local shadow effects. By focusing on these grassland ROIs, the analysis emphasizes observation-geometry-induced reflectance anisotropy rather than surface type differences.

### B. UAV Multispectral Data Acquisition

The multispectral dataset was acquired using a UAV platform equipped with a MicaSense RedEdge-MX Dual camera, which provides 10 discrete spectral bands spanning the visible to near-infrared range. The sensor adopts a frame-based imaging mode to capture multispectral images.

Data acquisition was conducted during a single flight mission under clear-sky conditions, resulting in several hundred multispectral frames. Due to the low flight altitude

This work was supported by an Internal Independent Research Project of the University of Information Engineering and by Heilongjiang Provincial Outstanding Youth Science Fund Project (Grant No. YQ2024F003).

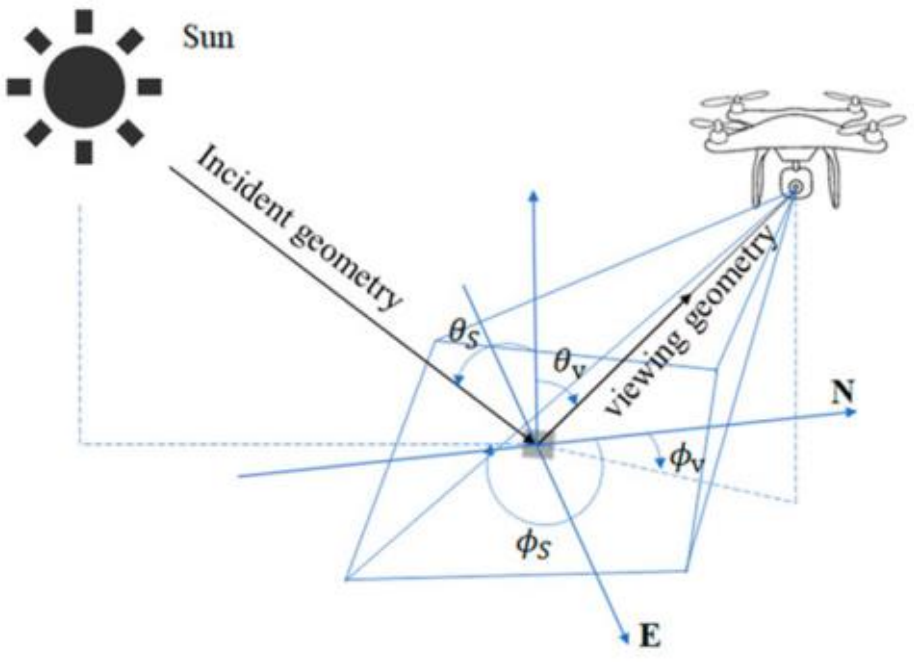

Fig. 2. Illustration of observation geometry.

and the wide field-of-view of the sensor, the acquired imagery naturally contains a wide range of viewing geometries, from near-nadir to relatively large off-nadir angles, even within the same flight.

## III. OBSERVATION GEOMETRY

Multi-angular reflectance observed in UAV multispectral imagery arises from variations in illumination and viewing geometry. Following standard BRDF geometry conventions, the observation geometry of each pixel is described by the solar zenith angle $\theta_s$, view zenith angle $\theta_v$, solar azimuth angle $\phi_s$, and viewing azimuth angle $\phi_v$, as illustrated in Fig. 2. The relative azimuth angle $\phi$ is defined as.

$$\phi = |\phi_v - \phi_s| \tag{1}$$

Let $\mathbf{n}$ denote the surface normal vector, $\mathbf{s}$ the solar illumination direction, and $\mathbf{v}$ the sensor viewing direction. The corresponding zenith angles are computed as:

$$\theta_s = \arccos(\mathbf{n}\cdot\mathbf{s}),\ \theta_v = \arccos(\mathbf{n}\cdot\mathbf{v}) \tag{2}$$

Under these definitions, surface reflectance in band λ can be expressed as a function of observation geometry:

$$R^{\lambda} = R^{\lambda}(\theta_s, \theta_v, \phi) \tag{3}$$

The directional dependence of reflectance can be interpreted under the BRDF framework. In remote sensing, BRDF is often represented using an approximate form such as a linear combination of geometry-dependent kernel functions:

$$R(\theta_s, \theta_v, \phi) \approx \sum_i a_i K_i(\theta_s, \theta_v, \phi) \tag{4}$$

Where $K_i$ denotes the kernel functions and $a_i$ represents the corresponding coefficients. In this study, we focus on observation-geometry-driven reflectance variations captured by UAV images. Since solar geometry varies only slightly during a single UAV flight, subsequent analysis primarily emphasizes reflectance patterns associated with $\theta_v$ and $\phi$.

## IV. METHOD

Fig. 3 illustrates the overall workflow of the proposed geometry-aware multi-angular observation extraction and analysis pipeline. The method consists of four main stages: (1) SFM-based reconstruction for refined camera parameters and orthomosaic generation, (2) ROI annotation on the orthomosaic, (3) geometry-aware observation extraction via reprojection to raw images, and (4) polar visualization of multi-angular reflectance.

### A. Data Preprocessing

Prior to multi-angular analysis, the UAV multispectral images were preprocessed to improve both radiometric and geometric consistency. Radiometric calibration was applied to convert raw DNs to reflectance. In addition, band-to-band registration was performed to correct potential misalignment among spectral bands, ensuring that reflectance measurements from different bands correspond to the same ground targets.

### B. SFM-based reconstruction

To obtain accurate imaging geometry for multi-angular analysis, a SFM based reconstruction is first performed using multi-view UAV imagery. The objective of SFM is to jointly estimate camera intrinsic parameters, camera poses, and the three-dimensional structure from image correspondences.

In the SfM framework, the projection of a 3D point $\mathbf{X}_j$ onto the image plane of the $i$ -th view can be expressed as

$$\mathbf{x}_{ij} : \mathbf{K}_i(\mathbf{R}_i\mathbf{X}_j + \mathbf{t}_i) \tag{5}$$

where $\mathbf{K}_i$ denotes the intrinsic matrix, and $\mathbf{R}_i$, $\mathbf{t}_i$ represent the rotation and translation of the camera pose, respectively.

After establishing multi-view correspondences, bundle adjustment is applied to refine camera poses and 3D points by minimizing the reprojection error:

$$\min_{\mathbf{R}_i,\mathbf{t}_i,\mathbf{K}_i} = \sum_{i,j}\left\| \mathbf{x}_{ij} - \pi\left(\mathbf{K}_i(\mathbf{R}_i\mathbf{X}_j + \mathbf{t}_i)\right)\right\|^2 \tag{6}$$

where $\pi(\cdot)$ denotes the projection operator.

For each view, SFM provides the refined camera intrinsics $\hat{\mathbf{K}}_i$ and extrinsics $\hat{\mathbf{R}}_i$ and $\hat{\mathbf{t}}_i$, used to establish geometric correspondence between the orthomosaic and raw sub-images.

### C. ROI annotation

To analyze reflectance anisotropy under controlled surface conditions, representative grassland ROIs were manually annotated on the orthomosaic using ENVI or ArcGIS. The

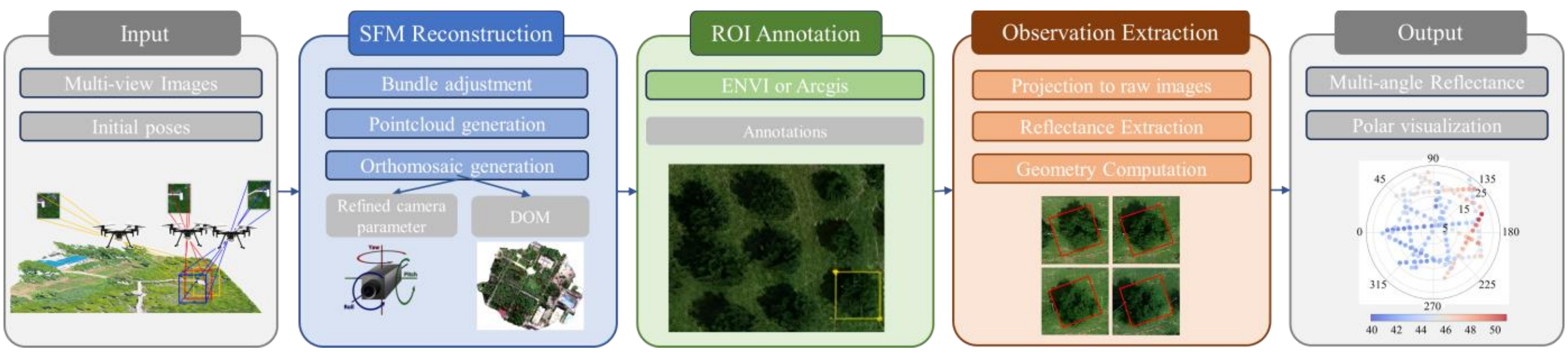

Fig. 3. Workflow of the proposed pipeline.

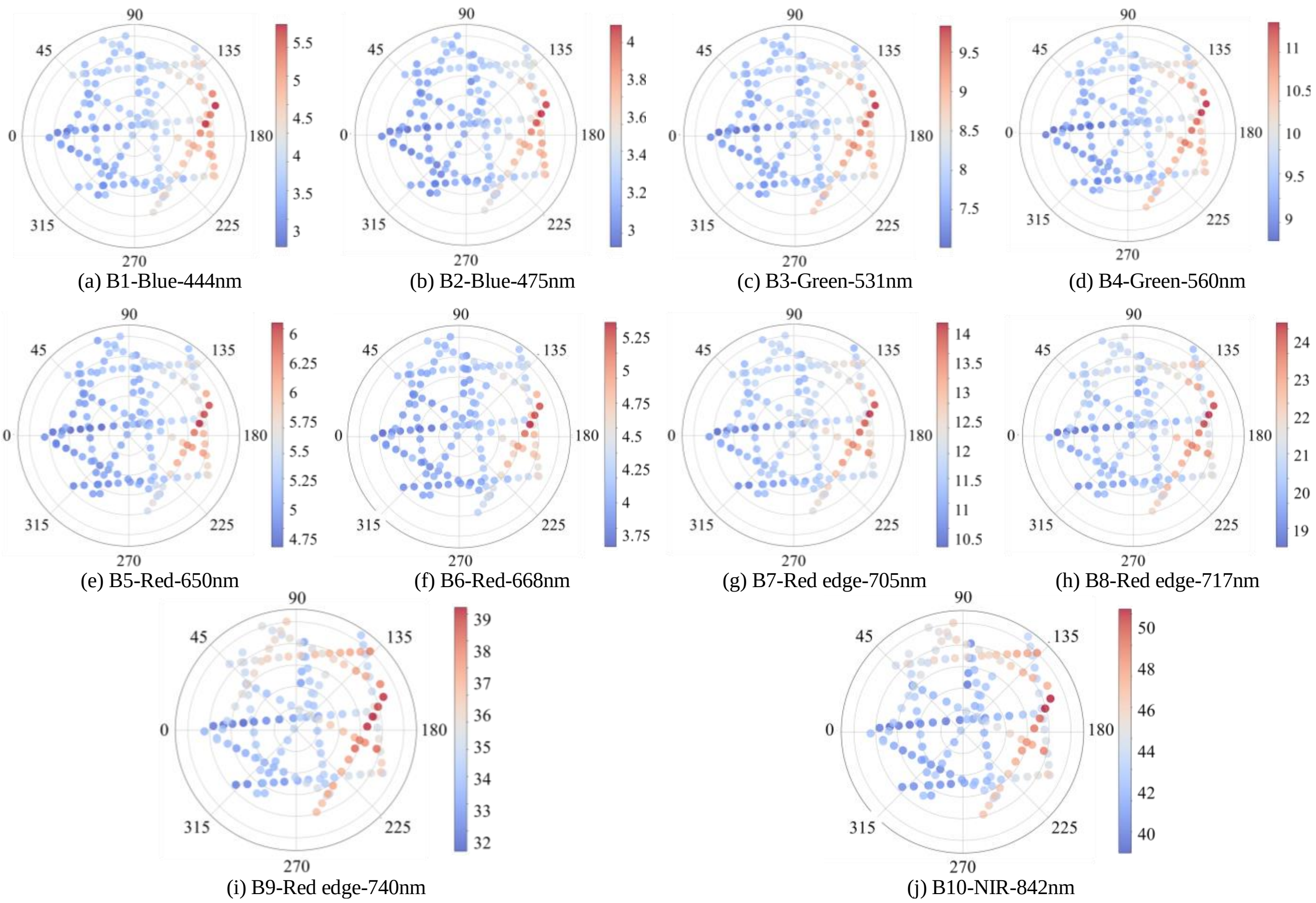
(a) B1-Blue-444nm (b) B2-Blue-475nm (c) B3-Green-531nm (d) B4-Green-560nm
(e) B5-Red-650nm (f) B6-Red-668nm (g) B7-Red edge-705nm (h) B8-Red edge-717nm
(i) B9-Red edge-740nm (j) B10-NIR-842nm

Fig. 5. Polar maps of multi-angular reflectance across ten spectral bands for the grassland region. The radial coordinate denotes the view zenith angle, the angular coordinate denotes the relative azimuth angle, and color indicates reflectance magnitude.

selected ROIs exhibit relatively homogeneous texture and spectral uniformity, which helps reduce the influence of land-cover heterogeneity and enables the analysis to focus on geometry-induced reflectance variations.

### *D. Observation Extraction*

Based on the refined camera intrinsics $\hat{\mathbf{K}}_i$ and extrinsics $\hat{\mathbf{R}}_i$ and $\hat{\mathbf{t}}_i$ obtained from the SFM reconstruction and the ROI annotations from Section IV.C, the image projection result on the $j$-th sub-image follows the perspective projection model:

$$\hat{\mathbf{x}}_j : \hat{\mathbf{K}}_j\left(\hat{\mathbf{R}}_j\mathbf{X}+\hat{\mathbf{t}}_j\right) \tag{7}$$

where $\hat{\mathbf{x}}_j$ denotes the pixel location in the raw sub-image. Using this geometric relationship, ROI pixels are sampled in each view and the corresponding band reflectance values are extracted, denoted as $R_j^{\lambda}$ for band $\lambda$.

For each valid reprojection, band-wise reflectance values were extracted from the corresponding sub-image. Meanwhile, observation geometry parameters $(\theta_s, \theta_v, \phi)$ were computed for each sample based on the refined camera pose and the illumination direction described in Section III.

Aggregating all ROI samples across available views forms a multi-angular observation set:

$$\pounds^{\lambda} = \left\{\left(\hat{\mathbf{x}}_j, R_j^{\lambda}\left(\theta_v, \phi\right)\right)\right\}_{j=1}^{N} \tag{8}$$

where $N$ is the number of valid views.

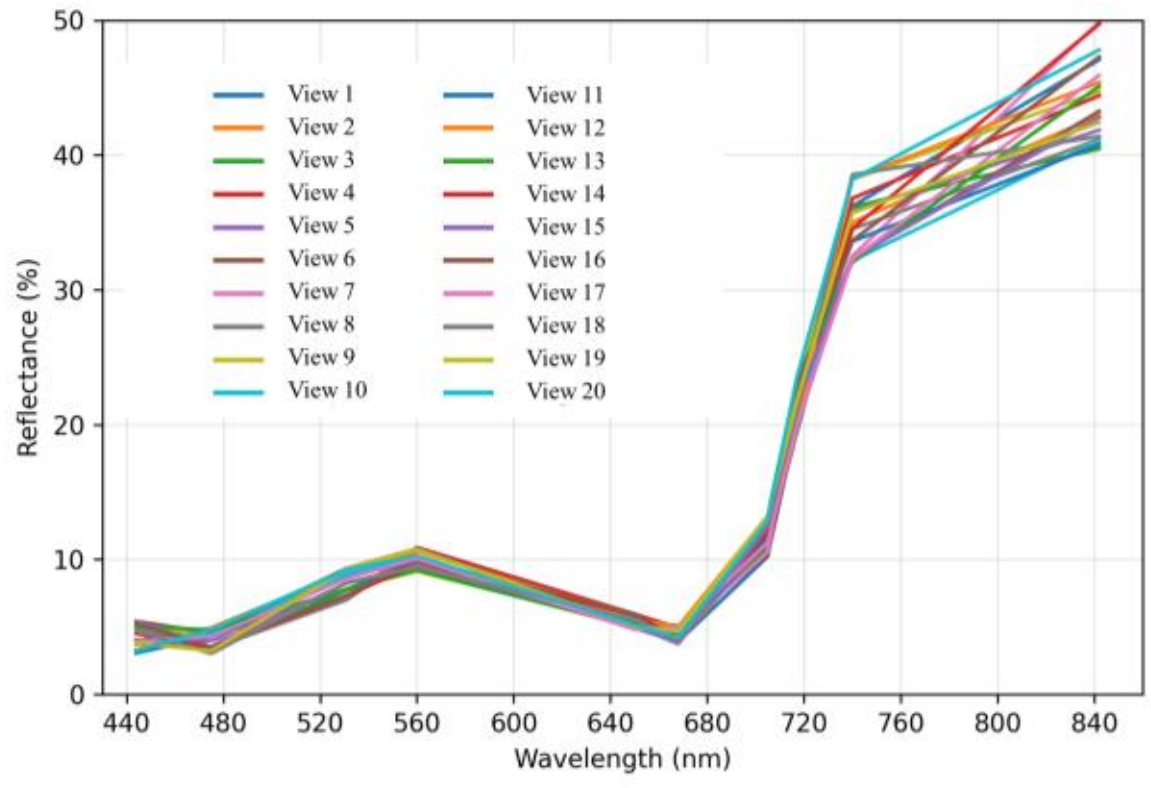

Fig. 4. Spectral reflectance curves of a grassland ROI observed from 20 viewing angles.

### *E. Polar Visualization of Multi-Angular Reflectance*

To provide an intuitive representation of observation-geometry-driven reflectance variations, polar-coordinate visualization was applied to each spectral band. In the polar plots, the radial coordinate represents the view zenith angle $\theta_v$, while the angular coordinate corresponds to the relative azimuth angle $\phi$. The reflectance in each band is encoded by the color intensity. By comparing polar plots across spectral bands, the spectral dependence of multi-angular anisotropy can be visually assessed under the BRDF framework.

## V. RESULT

Based on the proposed geometry-aware multi-angular observation extraction framework, the reflectance characteristics of the grassland ROI under varying observation geometries are analyzed. By combining refined camera parameters obtained from SfM reconstruction with ROI annotation in orthomosaic space, consistent geometric correspondence across multiple views is established, ensuring the comparability of reflectance observations acquired from different viewing angles.

Figure 4 illustrates the polar-coordinate distributions of multi-angular reflectance for different spectral bands under explicitly defined observation geometry. Systematic reflectance variations with respect to view zenith and relative azimuth angles are observed in all bands, indicating that the extracted reflectance changes are strongly associated with observation geometry rather than random noise or misregistration effects. This confirms the effectiveness of the geometry-aware extraction strategy in characterizing reflectance anisotropy.

From a spectral perspective, the red-edge and near-infrared bands (B7–B10) exhibit noticeable but relatively moderate angular dependence, with the maximum reflectance reaching approximately **119–137%** of the minimum value across viewing angles. In contrast, the visible bands show stronger sensitivity to observation geometry, with maximum-to-minimum ratios of approximately **106–227%**, highlighting more pronounced angular-induced fluctuations. It should be noted that visible-band reflectance over vegetation is typically low, and ratio-based metrics may therefore be more sensitive to small absolute changes. Nevertheless, these statistics highlight that observation geometry can introduce non-negligible radiometric variability across viewing directions, even for a homogeneous grassland target within a single UAV mission.

Figure 5 presents the spectral reflectance curves of the same grassland ROI observed from 20 viewing angles. Owing to the geometry-aware ROI back-projection and observation extraction, the spectral shapes remain consistent across all views, while the reflectance magnitudes vary systematically with viewing geometry. These results further demonstrate that the observed variability is primarily driven by geometry-dependent effects rather than intrinsic surface heterogeneity.

## VI. CONCLUSION

This study presents a UAV-based framework for multi-angular reflectance analysis using multispectral imagery. Refined camera parameters are obtained through SFM reconstruction and combined with ROI annotation in orthomosaic space to enable accurate extraction of reflectance observations for the same surface area under multiple viewing angles. Based on this framework, polar-coordinate visualization is employed to systematically analyze angular-dependent reflectance characteristics across spectral bands.

The experimental results demonstrate that pronounced angular effects exist even over relatively homogeneous grassland surfaces. Noticeable reflectance variations are observed across viewing angles in the red-edge and near-infrared bands, while the visible bands exhibit even stronger angular-induced fluctuations. These findings highlight the non-negligible influence of observation geometry in UAV-based multispectral reflectance analysis.

The proposed workflow provides a practical basis for organizing and analyzing multi-angular UAV observations. Future work will focus on integrating quantitative BRDF modeling and multi-temporal observations to further improve the reliability of UAV multispectral data for surface parameter retrieval and fine-scale remote sensing applications..